\pdfoutput=1

\documentclass[11pt]{article}

\usepackage{acl}

\usepackage{times}
\usepackage{latexsym}

\usepackage[T1]{fontenc}

\usepackage[utf8]{inputenc}

\usepackage{microtype}

\usepackage{graphicx}
\usepackage{tikz}   
\usepackage{multirow}
\usepackage[utf8]{inputenc} 
\usepackage[T1]{fontenc}    
\usepackage{hyperref}       
\usepackage{url}            
\usepackage{booktabs}       
\usepackage{amsfonts}       
\usepackage{nicefrac}       
\usepackage{microtype}      
\usepackage{times}
\usepackage{latexsym}
\usepackage{amsmath}
\usepackage{graphicx}
\usepackage{multirow}
\usepackage{enumitem}
\usepackage{colortbl}
\usepackage{xcolor}

\usetikzlibrary{shapes.geometric, arrows}
\tikzstyle{io} = [rectangle, rounded corners, minimum width=1.5cm, minimum height=.8cm,text centered, draw=black, fill=red!30]
\tikzstyle{process} = [rectangle, minimum width=1.5cm, minimum height=.8cm, text centered, draw=black, fill=orange!30]
\tikzstyle{arrow} = [thick,->,>=stealth]

\title{TransPOS: Transformers for Consolidating Different POS Tagset Datasets}

\author{Alex Li$^1$, Ilyas Bankole-Hameed$^1$, Ranadeep Singh$^1$, Gabriel Shen Han Ng$^1$, Akshat Gupta$^2$\\
$^1$Carnegie Mellon University, $^2$JPMorgan AI Research, New York, USA\\
\texttt{\{alexli2, ibankole, ranadees, hsng\}@andrew.cmu.edu}\\
\texttt{akshat.x.gupta@jpmorgan.com}
}

\begin{document}
\maketitle

\begin{abstract}
In hope of expanding training data, researchers often want to merge two or more datasets that are created using different labeling schemes. This paper considers two datasets that label part-of-speech (POS) tags under different tagging schemes and leverage the supervised labels of one dataset to help generate labels for the other dataset. This paper further discusses the theoretical difficulties of this approach and proposes a novel supervised architecture employing Transformers to tackle the problem of consolidating two completely disjoint datasets. The results diverge from initial expectations and discourage exploration into the use of disjoint labels to consolidate datasets with different labels. 
\end{abstract}

\section{Introduction}
There has been an explosion in the availability and variety of labeled datasets in almost every domain. Unfortunately, Artificial Intelligence (AI) practitioners and researchers often find themselves unable to make use of labeled datasets for tasks related but not identical to their tasks. This is primarily due to different labeling schemes where a trivial mapping to merge the datasets into one larger dataset does not exist. In this paper, we explore the possibility of consolidating datasets that were curated for the same task with different labeling schemes. To make this easy to apply to any pair of datasets, we consider a very interesting scenario in which we attempt to make a model that can understand both datasets without ever actually seeing any examples that have labels from both of them. 

There are several domains and application areas to which our technique can be applied to, and frankly might be the only option. For example: When creating a data set to detect people, objects, and vehicles in an urban environment, we may want to supplement our existing data set with the popular Cityscapes dataset \citep{cityscapes}, but struggle to directly apply those labels to the merged dataset due to a few minor differences in the label scheme, such as a smaller or larger label set. There could also be some information partially correlated with the existing dataset's labels; perhaps in our dataset we have to distinguish between standing and sitting people. Cityscapes does not distinguish between these, so is it possible to use the label information (about where humans are) to get high-quality segmentation under our desired labeling scheme?

The focus of this paper comes from part-of-speech (POS) tagging. Although some tags are common to all datasets, different datasets may have different conventions for how to deal with more uncommon parts of speech, like modal verbs, particles, or even when to treat something as a noun. These problems are exacerbated in informal contexts. We provide a novel design for a supervised model that can translate labels from one dataset into another labeled dataset \textit{without requiring any shared examples}. After analyzing results, we reconsider the situations under which it is possible to squeeze out extra performance from these labels, and show that it is  unlikely for any kind of architecture to use label information to perform better than an equivalent model that does not, unless the architecture has access to shared examples or metadata about the meaning of the labels.

\subsection{Related Work}
The problem of dissimilar POS tagsets has historically been approached in two significant ways: 
\begin{enumerate}
\item Supervised Learning: \citep{Shen07guidedlearning} proposed a supervised POS tagger with 97.3\% accuracy for the English language;
\item Create  Dictionary Mapping: \citep{DBLP:journals/corr/abs-1104-2086} proposed a Universal POS tagset to map 25 different treebank tagsets to 12 universal POS tags. 
\end{enumerate} 
There has also been a significant amount of progress in creating POS tags for languages other than English leveraging both supervised and unsupervised methods \citep{37071}. 

Our work can be seen as a type of Multitask Learning \citep{DBLP:books/sp/98/Caruana98} as we are learning from two related datasets that have been labeled independently and differently. A common technique is to create a model for each task \citep{collobert2008unified}, and enforce weight sharing between their lower layers to allow shared low-level domain knowledge.
A key distinction between our methodology and Multitask Learning is that our test time goal also makes use of labels from the other task. We use the actual predictions of the model rather than the more common idea of using the predicted logits or encoded representation from a previous layer.

This problem can also be considered as a type of Domain Adaptation Technique. However, many domain adaptation algorithms (\citep{Frustratingly_Easy_Domain_Adaptation}) assume some shared examples between the source and target domains, so we cannot apply it in our case. Those algorithms that do not make this assumption have never to our knowlege tried to use the labels that are in the target distribution but are not the source distribution labels.

\section{Setting}
Let $\Sigma$ be the set of unicode characters.
In our setting, we have two datasets that map from the space of sentences of unicode characters $X=\bigcup \Sigma^n$ to part-of-speech tags. However, the two datasets use different labeling schemes: the first may use the standard Universal POS tagset $Y$ while the second uses a proprietary POS tagset $Z$. Each sentence in the first dataset has a label for each word in $\bigcup_{n=0}^\infty Y=\mathbf{Y}$, while each sentence in the second has a label in $\bigcup_{n=0}^\infty Z=\mathbf{Z}.$

Then we can name the two datasets as $\mathcal{D}_Y =\{(x_y^{(i)},y^{(i)}) \in (X,\mathbf{Y})\}_i$ and $\mathcal{D}_Z = \{(x_z^{(i)}, z^{(i)})\in (X, \mathbf{Z})\}_i$. Presumably, $\mathbf{Y}$ and $\mathbf{Z}$ are very highly correlated, since they are both POS tags for a sentence, just defined with slightly different rules. We would like to expand the dataset $\mathcal{D}_Y$ to include the sentences and labels of $\mathcal{D}_Z$, but unfortunately the labels are incompatible. However, we expect that we can still get useful information from the labels $\mathbf{Z}$. Therefore, our goal is to build a predictor function $f_{Y}: (X,\mathbf{Z})\to \mathbf{Y}$ that combines both the text and the information of the annotated $\mathcal{D}_Z$ to predict what the translated label would be in the tagset $\mathbf{Y}$. Similarly, we consider the construction of $f_{Z}:(X,\mathbf{Y})\to \mathbf{Z}$.

Here are the two obvious baselines that could be used to construct $f_{Y}$:
\begin{enumerate}
\item \textbf{Direct Map} We could use domain knowledge to directly design a mapping from each label $Z\to Y$. If $|Z|< |Y|$, $Y$ is not a deterministic predictor of $Z$ and this introduces noise into the system. If $|Z| > |Y|$, converting to $Y$ will result in a loss of information.
\item \textbf{Supervised Model} We could train a model on $\mathcal{D}_Y$ to build a function $X\to \mathbf{Y}$. 
\end{enumerate}
Note that while the second baseline is trained with data, the first baseline is completely based on human understanding of the relationship between labels. Thus, while we can naturally train a model to match the performance of the \textbf{Supervised Model}, it is much less obvious how we can train a model to gain the performance advantage given by the \textbf{Direct Map} method.

Now we consider the design of our model intended to use information about both X and $\mathbf{Z}$ to perform better than either approach.

\section{Model}
In our method, we will transform our input $X$ into an embedding space $E$ using a transformer's encoder $Enc:X\to E$ and two GRU decoder functions, one for each type of label $\mathbf{Y}$ and $\mathbf{Z}$. $D_{Y}:(E, \mathbf{Z})\to \mathbf{Y}$ and $D_{Z}:(E, \mathbf{Y}) \to \mathbf{Z}$. Then to infer a label Z for a given training sample $(x_y, y)$, we can compute $D_{Z}(Enc(x_y), y).$ See Figure \ref{fig:eval_mapper} for a visualization.

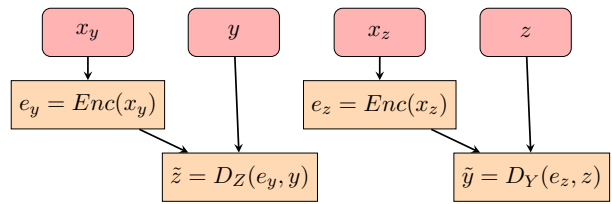
\begin{figure}[!ht]
    \centering
\scalebox{.8}{\begin{tikzpicture}[node distance=1.2cm]
\node (XY) [io] {$x_y$};
\node (Y) [io, right of=XY, xshift=1.2cm] {$y$};
\node (XZ) [io, right of=Y, xshift=1.2cm] {$x_z$};
\node (Z) [io, right of=XZ, xshift=1.2cm] {$z$};
\node (EY)[process, below of=XY]{$e_y = Enc(x_y)$};
\node (EZ)[process, below of=XZ]{$e_z = Enc(x_z)$};
\node (yyhead) [process, below of=EY,xshift=2.5cm]{$\tilde{z} = D_Z(e_y, y)$};
\node (zzhead) [process, below of=EZ,xshift=2.5cm]{$\tilde{y} = D_Y(e_z, z)$};
\draw [arrow] (XY) -- (EY);
\draw [arrow] (XZ) -- (EZ);
\draw [arrow] (Y) -- (yyhead);
\draw [arrow] (Z) -- (zzhead);
\draw [arrow] (EY) -- (yyhead);
\draw [arrow] (EZ) -- (zzhead);
\end{tikzpicture}}
    \caption{When evaluating the model, we use $y$ and $z$ as inputs!}
    \label{fig:eval_mapper}
\end{figure}
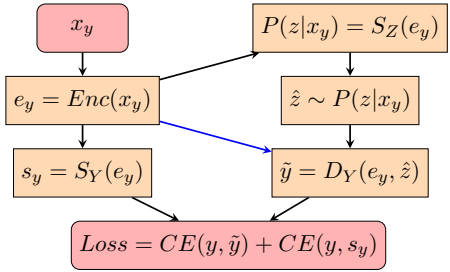
\begin{figure}[!ht]
    \centering
\scalebox{.8}{\begin{tikzpicture}[node distance=1.2cm]
\node (XY) [io, xshift=2.0cm] {$x_y$};
\node (PY) [process, right of=XY, xshift=3.2cm] {$P(z|x_y) = S_Z(e_y)$};
\node (EY)[process, below of=XY]{$e_y = Enc(x_y)$};
\node (PYY)[process, below of=EY]{$s_{y} = S_Y(e_y)$};
\node (yzhead) [process, right of=EY, xshift=3.2cm]{$\hat{z} \sim P(z|x_y)$};
\node (yyhead) [process, below of=yzhead]{$\tilde{y} = D_Y(e_y, \hat{z})$};
\node (output) [io, below of=yyhead,xshift=-2.0cm]{$Loss=CE(y, \tilde{y}) + CE(y, s_y) $};
\draw [arrow] (XY) -- (EY);
\draw [arrow] (PY) -- (yzhead);
\draw [arrow] (EY) -- (PY);
\draw [arrow] (EY) -- (PYY);
\draw [arrow, blue] (EY) --  (yyhead);
\draw [arrow] (yzhead) -- (yyhead);
\draw [arrow] (PYY) -- (output);
\draw [arrow] (yyhead) -- (output);
\end{tikzpicture}}
    \caption{To train the model with the $Y$ dataset, we simulate having Z labels by sampling from the logits of a normal supervised model $S_Z$. At the same time, we train a model $S_Y$ for the other dataset. Heavy dropout is applied at the location of the blue arrow.}
    \label{fig:training_mapper}
\end{figure}

However, the setup used for validation will not work for training the model. Ideally, we would like to make a loss function that penalizes the predicted value of $z$ from being far from the true $z$ corresponding to $x_y$, but we do not have any access to the true z! We only have pairs $(x, y)$ and $(x, z)$, not $(x, y, z)$. One way to fix this is to first predict $\hat{z}$ using $(x,y)$ and then use that as a surrogate for true $z$. To do this, we can create a supervised model $S_Z$ that takes in the encoded variables $Enc(x_y)$ and outputs a prediction for the label $z$, which we can then input into the decoder $D_Z.$ However, the careful reader will notice a flaw in this strategy: While $D_Z$ takes the input as one-hot labels at inference time, it takes input as logits at training time. To reconcile this difference, we treat the softmax of the predicted logits as a probability distribution, from which we sample our true predicted label $\hat{z}$.

The entire training process with the inputs $\mathcal{D}_Y$ is shown in Figure $\ref{fig:training_mapper}$, and the model for $\mathcal{D}_Z$ is made the same way, but with the y / z inputs flipped. In our implementation, each mini-batch contains some examples from both datasets. To reduce the complexity of the model, we use the same base encoder model weights for $S_Y, S_Z, D_Y,$ and $D_Z$, though in principal they could be different or only partially shared.

The inquisitive reader may wonder why we use $S_Z$ to predict labels $z$ instead of reusing labels $D_Z$ along with ground truth labels $y$.
In this case, we will have given the label that we want to predict as an input to the model, and so the model can simply learn to predict the input! For example, suppose that in our model, rather than sampling $\hat{z} \sim S_Z(e_y)$, we reused the decoder weights to get $\hat{z} := D_Z(e_y, y)$, then predicted $\tilde{y} = D_Y(e_y, \hat{z})$ (and similarly on the other side). Then, the model can simply learn to ignore the first argument of $D_Y$ and $D_Z$ and instead learn that $D_Y(-,z)$ and $D_Z(-,y)$ are inverses to each other. In this setup, it will perfectly predict all the training data, but it will be completely useless in practice. We actually tried this setup and found that the model would actually achieve performance competitive with the supervised model for a few epochs (perhaps due to regularization like dropout), but after training long enough, it learns the cheat and arrives at 0 training error and very high validation error.
Now, during training, $y$ and $z$ are completely derived from $x$. So, in principle, $D_Y$ and $D_Z$ may learn to ignore noisy outputs $y$ and $z$ and make predictions based solely on $x$. To prevent this, we enforce a very heavy dropout of $0.85$ on the first term before passing it as input.

The model and training code can be found in out Github repository\footnote{\url{https://github.com/Alex7Li/TransPOS}} in the footnote.

\section{Datasets}
\begin{table}[]
    \centering
    \begin{tabular}{|c|c|}
     \hline
    \textbf{Text} & \textbf{Ark Label} \\
    \hline
    New & Adjective \\
    FC & Proper noun\\
    Menu & Proper noun\\
    Utility & Proper noun\\
    2.0 & numeral\\
    \#apple & Proper noun \\
    http://t.co/VftFt2c & URL or email address \\
     \hline
     \hline
    \textbf{Text} & \textbf{Tweebank Label}\\
    \hline
    @USER2082 & A \\
    good & ADJ \\
    night & NOUN \\
    I & PRON \\
    Love & VERB \\
    You & PRON \\
    :) & SYM \\
    http://t.co/VftFt2c & U \\
     \hline
    \end{tabular}

    \caption{Example tweets from Ark and Tweebank}
    \label{tab:my_label}
\end{table}

In this project, we consider two datasets:
\begin{enumerate}
    \item ARK-Twitter \citet{brenocon2011}, which contains 34k tokens from tweets sampled primarily on Oct 27, 2010.
    \item Tweebank dataset \citet{YijiaLiu2018} which maintains 840 tweets from Tweebank v1, 2500 examples from twitter stream from February 2016 to July 2016.
\end{enumerate}
The Tweebank dataset used UD annotation conventions, while the ARK data set used the Stanford POS Tagger trained in WSJ.

However; these two datasets have a data contamination problem: there are 210 identical shared tweets. In our case, however, this served as the perfect validation set for our model.

\begin{figure}[]
    \centering
    \includegraphics[trim={1cm 0cm 3cm 0cm},clip, width=.5\textwidth]{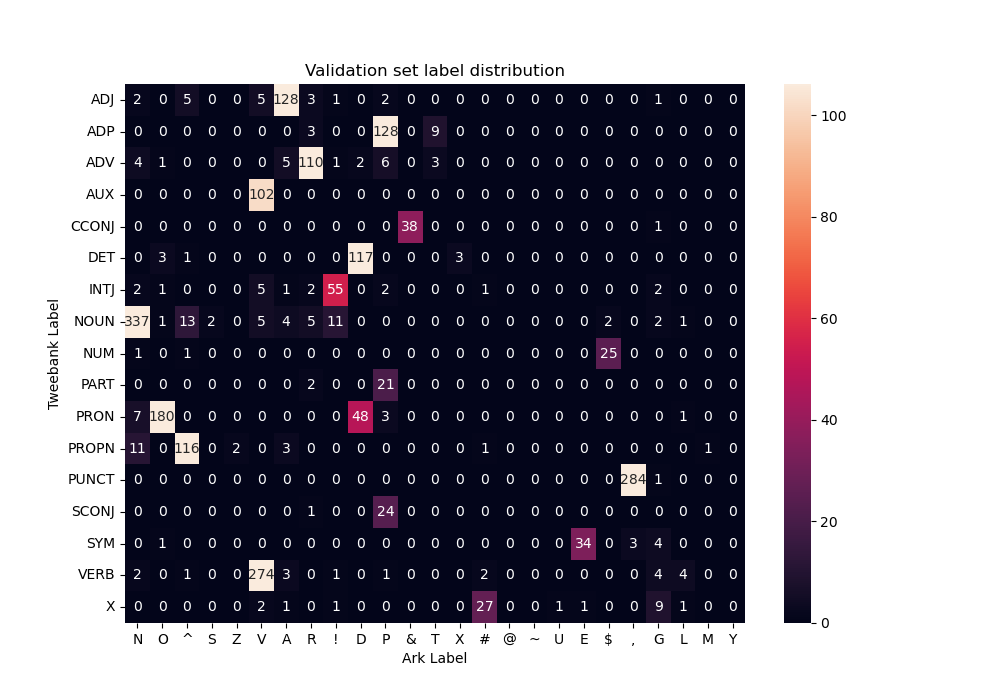}
    \caption{Distribution of validation labels.\\}
    \label{fig:valdiation_distribution}
\end{figure}

Looking at the distribution of the labels in this validation set (Figure \ref{fig:valdiation_distribution}), we see that the ambiguity between the meanings of the labels will limit the performance of a direct mapping.

\section{Experiments}

\begin{table}[ht]
    \centering
    \begin{tabular}{|c|c|c|}
    	\hline
        \textbf{GPT-2} & \textbf{Tweebank Acc} & \textbf{Ark Acc} \\
        \hline
        supervised model & \textbf{89.53}\% & 89.92\% \\
        our model & \textbf{89.53}\% & \textbf{90.17}\% \\
        supervisor only & 86.96\% & 88.29\% \\
        no label input & 88.09\% & 88.59\% \\
        \hline
        \hline
       \textbf{ Bertweet-large} & \textbf{Tweebank Acc} & \textbf{Ark Acc} \\
        \hline
        supervised model & \textbf{94.31}\% & 95.02\% \\
        our model & 94.26\% & 94.97\% \\
        supervisor only & \textbf{94.31}\% & \textbf{95.09}\% \\
        no label input & 94.22\% & 94.97\% \\
        \hline
        direct map & 88.31\% & 89.97\% \\
        \hline
    \end{tabular}
    \caption{Accuracy (Acc) with Bertweet-large model baseline}
    \label{tab:accuracies}
\end{table}
The first baseline was created by making a `direct map' between the labels. We looked at the validation set and chose the map that gave the highest possible score.

The second 'supervised model' baseline was a normal transformer model; we train on the train split of one dataset and evaluate on the validation split of that same dataset.

Then, for `our model', we trained with the described architecture, using the transformers BERTweet (\citet{Bertweet}), and GPT-2 (\citet{gpt2})
 a GPT-2 model and a Bertweet model with the described architecture with dropout $.85$. After the training was complete, we evaluated the accuracy using the method described above to compute our model accuracy.
 
To see if our model was really learning from the y labels, we used of the supervised model heads $S_{Y} \circ Enc$ and $S_{Z} \circ Enc$ to get the `supervisor only' accuracy. This architecture is exactly the same as the baseline, but differences arise in the accuracy because the training process is not the same (in particular, there is very high $x$ dropout).

Finally, we considered the accuracy of the full pipeline when there with `no label input': instead of providing the $z$ labels for the first dataset while predicting the $y$ labels of the second, we just took the $z$ labels that the model predicted and sampled from that distribution as we do at training time.

\section{Results}

\begin{figure}[!ht]
    \centering
    \includegraphics[trim={1cm 0cm 3cm 0cm}, width=0.47\textwidth]{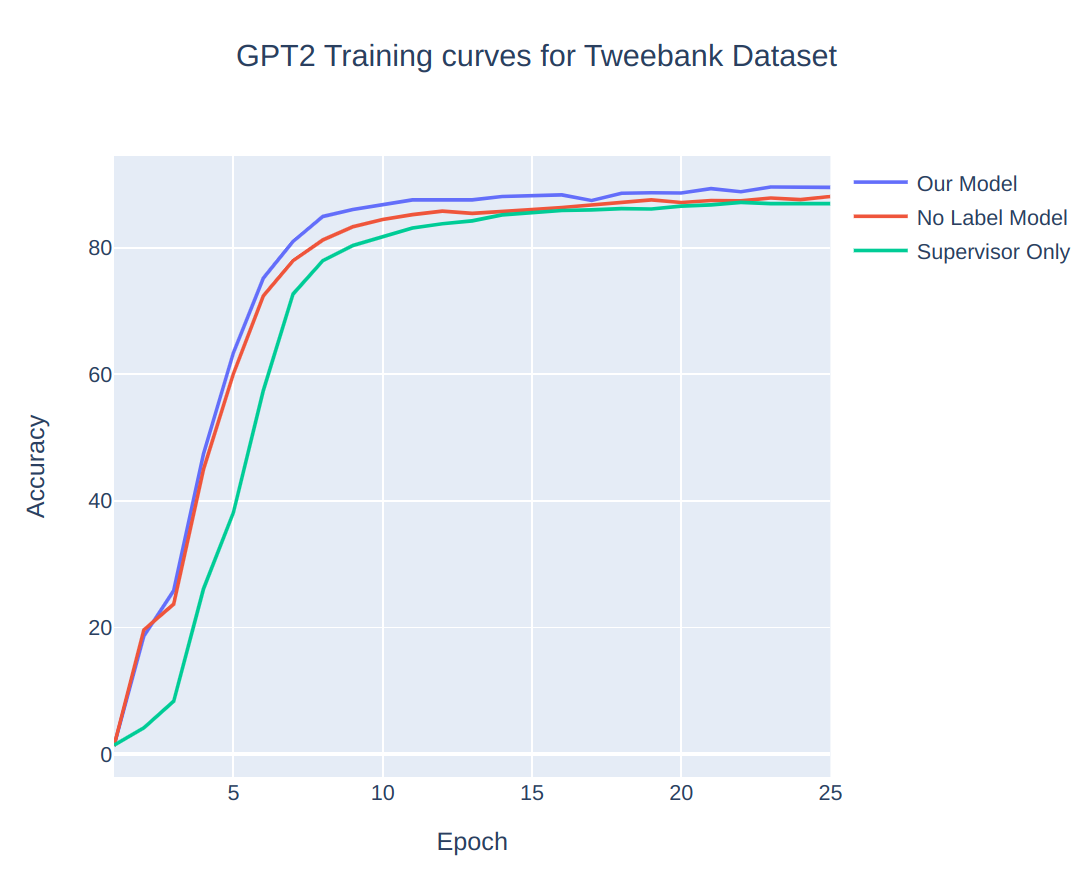}
    \caption{GPT-2 Validation accuracy}
    \label{fig:gpt2_twee_training_curve}
\end{figure}

We trained the model with both the GPT-2 as our encoder and the Bertweet model as our encoder. All models were trained for 25 epochs, and we report the accuracy at the final epoch in Table \ref{tab:accuracies}.

In both cases, our accuracy does not exceed the baseline. Although the Bertweet model appears to gain nothing from the $z$ labels, the GPT-2 model appears to be using the $z$ labels to effectively improve performance as indicated in Figure \ref{fig:gpt2_twee_training_curve}. 
Since the score of the model improves when we give it the z labels, we can say that it is actually learning to use the joint probability distribution of the $y$ and $z$ labels.



Our approach can only be useful when the correlation between the labels of both datasets provides information that the correlation between the encoder output and the true labels does not. One possible explanation for our inability to beat the baseline model is because the label information was not sufficient or because the baseline model was already too strong. However, using the weaker GPT-2 model as a baseline did not show any improvement. 

Although there are a multitude of different ideas for model designs that use the $y$ labels, it is important to first try and understand why this model struggled in this regime. From our results, it seems it will be difficult to design an architecture that can effectively learn from the label information of another dataset without using any shared examples.

To emphasize that this problem will be hard for any architecture, let us consider a toy example of this problem where we no longer have any $X$ data and are just given a set of $\mathbf{Y}$ POS tokens and $\mathbf{Z}$ POS tokens. In this case, the the $y$ and $z$ labels are still very correlated, but since it is impossible for a model to predict the price from an integer id, our model will not be able to learn about and make use of the high covariance between labels. As we have given the model two unrelated sets of labels, no matter what model you use, it will be impossible to relate them with anything other than the statistical properties of the $\mathbf{Y}$ and $\mathbf{Z}$ distributions. This does not seem too informative in general, since it will be difficult to find the correct relationship between two sets with no shared examples, though the fact that POS is a multi-label prediction problem means that you might be able to get a bit out of it. Still, even trying to make the label distributions similar is not easy as the labels are not in the same space. 

In our model, we consider pairs of predicted $y$ and true $z$ data, which ultimately cannot give any more information than the already known relationship between encoder outputs and true labels. The hope was that replacing the predicted label $y$ with the true label $y$ would allow for a final gain in accuracy, but that was not the case in our experiments. There is a difficult tension to balance: When trusting the predicted $y$ label too much, the decoder will not be able to perform well on the training dataset because the predicted label is often wrong. But when we do not trust the label, we cannot do well at evaluation time.

The toy example indicates that the only other way to gain new information would be relating the statistical properties of the distribution. However, it is not clear how to learn this information or how helpful it would be. Therefore, using label information for a separate dataset appears very unlikely to improve performance.

A counterpoint to this argument is the performance gap between \textit{our model} and the \textit{no label input} model. This is especially clear in the early stages of GPT-2 training. In Figure ~\ref{fig:gpt2_twee_training_curve}, we plot the three accuracies when training GPT-2. Here, using the ground truth labels for the y dataset gives a better score on the z dataset than using the model predictions for y. Thus, it seems that we can conclude that it is possible for the model to learn the joint distribution $P(Y,Z)$ and use that information effectively. However, the problem is that the only information about $P(Y,Z)$ that the model is capable of learning is what can be deduced from $P(X,Y)$ and $P(X,Z)$. In trying to predict $Z$, it can really only use the information that was learned from $P(X, Z)$, which is already contained in any normal supervised model.
The fact that the model performance never surpasses the supervised model is evidence toward the argument that the replacement policy will not help to improve model performance in general.

\section{Conclusion}

The task of consolidating datasets with different labels and no shared examples is a hard problem. The experiments did not provide any improvement over the baseline of only using the $x$ variables. This was surprising, as the correlation between the $y$ and $z$ labels is quite large. However, this may be due to an intrinsic difficulty with the setting (no shared examples) rather than the model design.

\section{Future Work}
Future work of consolidating datasets without shared examples should focus on using semi-supervised learning with other $x$ labels or supporting the other dataset labels with metadata.

Another possible direction would be to use the architecture in this paper together with a subset of shared examples between the datasets. Our approach can be easily modified to deal with labels that are sometimes missing instead of all the time. Such a modification could shine in (potentially multi-label) environments with that require frequent missing value imputation.

\bibliography{TransPOS}
\newpage

\end{document}